\documentclass[10pt]{article}
\usepackage[colorlinks=true]{hyperref}
\usepackage{amsmath, amssymb,xspace}
\usepackage{amsfonts}
\usepackage[T1]{fontenc}
\usepackage{algorithm}
\usepackage{algorithmic}
\usepackage{caption}
\usepackage{subcaption}
\usepackage{multirow}
\usepackage{tikz}
\usepackage{pgfplots}
\usepackage{wrapfig}
\usepackage[inline]{enumitem}

\usepackage{graphicx,color,stackrel,xcolor}
\usepackage{color}

\definecolor{darkred}{RGB}{159,67,67}
\definecolor{lightgreen}{RGB}{100,255,100}

\newcommand{\lard}{{\sc lard}\xspace}

\definecolor{turquoise}{rgb}{.45,.93,.90}
\definecolor{redish}{rgb}{.95,.40,.45}
\definecolor{greenish}{rgb}{.63,.89,.60}

\newtheorem{definition}{Definition}
\newtheorem{mytask}{Task}

\usepackage[verbose,a4paper,lmargin=2.5cm,rmargin=2.5cm,tmargin=2.5cm,bmargin=3cm]{geometry}

\begin{document}
\title{LARD -- Landing Approach Runway Detection --\\
Dataset for Vision Based Landing}

\author{M\'elanie Ducoffe $^1$
\and
Maxime Carrere$^2$ 
\and
Léo Féliers$^3$
\and
Adrien Gauffriau$^1$
\and
Vincent Mussot$^4$
\and
Claire Pagetti$^3$
\and
Thierry Sammour$^1$
\and
$^1$ Airbus, France -- $^2$ Scalian, France -- $^3$ ONERA, France --
$^4$ IRT Saint Exupéry, France}
\maketitle
\begin{abstract}
As the interest in autonomous systems continues to grow, one of
the major challenges is collecting sufficient and representative
real-world data. Despite the strong practical and commercial interest in
autonomous landing systems in the aerospace
field, there is a lack of open-source datasets of aerial images. To
address this issue, we present a dataset -\lard- of high-quality aerial images for the task of runway detection during approach and landing phases. 
Most of the dataset is composed of synthetic images but
we also provide manually labelled images from real landing footages, to extend the detection task to a more realistic setting.
In addition, we offer the generator which can produce such synthetic front-view images and enables automatic annotation of the runway corners through geometric transformations.  
This dataset paves the way for further research such as the analysis of dataset quality or the development of models to cope with the detection tasks.
Find data, code and more up-to-date information
at \url{https://github.com/deel-ai/LARD}

\end{abstract} 

\section{Introduction}
Recent advances in Artificial Intelligence has made AI-based systems attractive to various fields, including transportation. However, in the aeronautic domain where these algorithms could increase autonomy,
AI breakthroughs are slow to reach the market, partly due to the lack of dedicated datasets.

\subsection{Context -- why vision based landing}
Increasing the level of autonomy of aircraft will ease the flying in case of pilots cognitive load and would therefore improve the safety in civil aviation. 
Today, aircraft already have on-board functionalities, that allow for complete automatic landings. But it has a huge impact on airport operations (number of landing per hour), high installation and maintenance costs. Thus the actual solution can only be used in bad weather conditions with impact on the landing rate. When the visibility is correct, the final phase still requires the pilot to see the runway at a specific distance.

In a future where it is envisaged to fly with only one pilot on board, a single pilot may not be in capacity to assume all tasks required during the landing phase (especially the final ones). Thus, it is required for the aircraft to be capable of performing landings without impacting the airport operations by merging information from several systems.
Considering recent advances in both computer vision and embedded hardware platforms make vision-based algorithms a potential direction in participating to the guidance and navigation during the landing stage.
A vision-based landing system will have to detect very distant to close runways on high resolution images. 
There are thus three challenges to tackle: propose 1) an efficient vision-based algorithm for the detection of runways of 2) very variable groundtruth sizes with 3) low execution time. 

\subsection{Importance of datasets}
To design deep learning vision-based landing systems, one mandatory step is to define datasets.
Indeed, 
high-quality, large-scale datasets are crucial for autonomous driving research. In the recent years, 
there have been an increasing number of efforts in that direction: releasing datasets to the community~\cite{geyer2020a2d2}, open-source simulators that allow us to generate scenarios and images to enable AI for autonomous cars like the CARLA simulator~\cite{dosovitskiy2017carla} or playground environment~\cite{paull2017duckietown} among other.
Regrettably, whether it is about real image retrieval or open-source simulators, the field of AI for aeronautics flights is way behind. It is very difficult today to retrieve images in flight and furthermore getting their metadata. We may also rely on simulator to easily generate images with associated metadata, but up to our knowledge, there is no good quality open-source simulator that could be used for image generation.
As a consequence, this lack of dataset impedes the widespread of AI product for the aeronautical field. 

\subsection{Contributions}
In this work, we want to close the gap and offer an open source dataset for runway detection to the community. 
We started by clearly defining the task at hand: detecting a runway in an image taken during the landing phase of an aircraft. In order to determine which images could potentially be taken during landing, we established a formal definition of the generic landing approach cone.
Based on this definition, we developed a strategy for constructing the dataset to encompass a wide range of scenarios.
Our approach involves 1) blending full approaches 
with a large or limited number of approach images, and 
2) taking into account various airports with diverse environments such as urban or rural areas, as well as different times of day when the images were captured.

\begin{figure}[hbt]
    \centering
    \begin{subfigure}[t]{.49\linewidth}
        \centering
        \includegraphics[width=\textwidth]{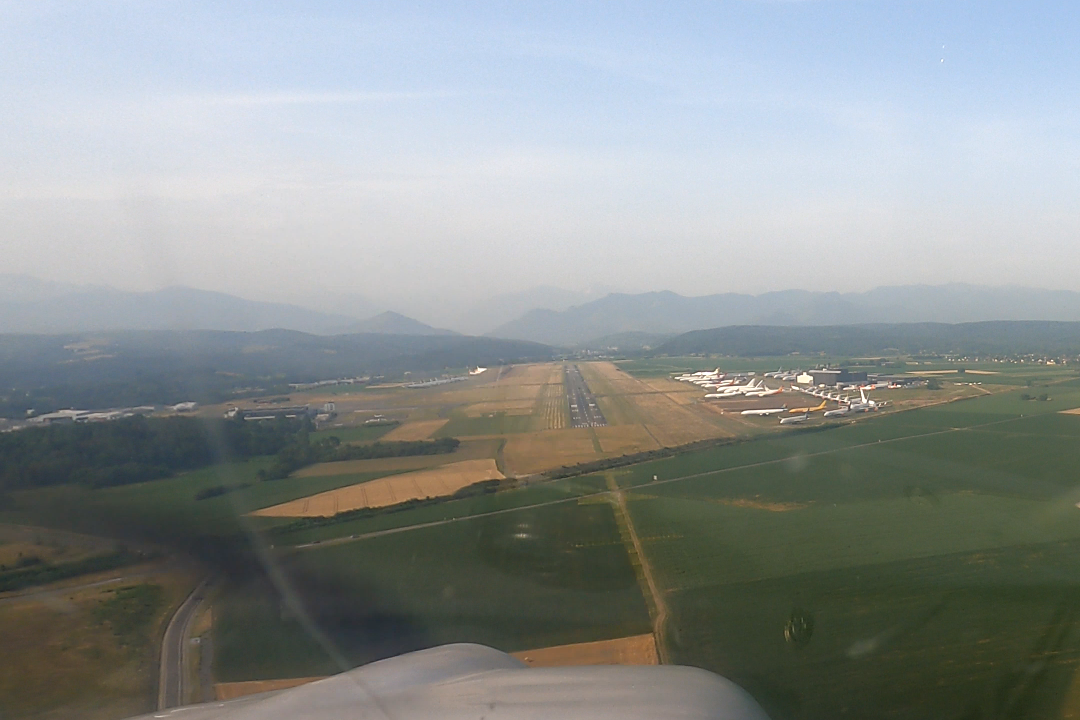}
    \end{subfigure}
    \hfill
    \begin{subfigure}[t]{.49\linewidth}
        \centering
        \includegraphics[width=\textwidth]{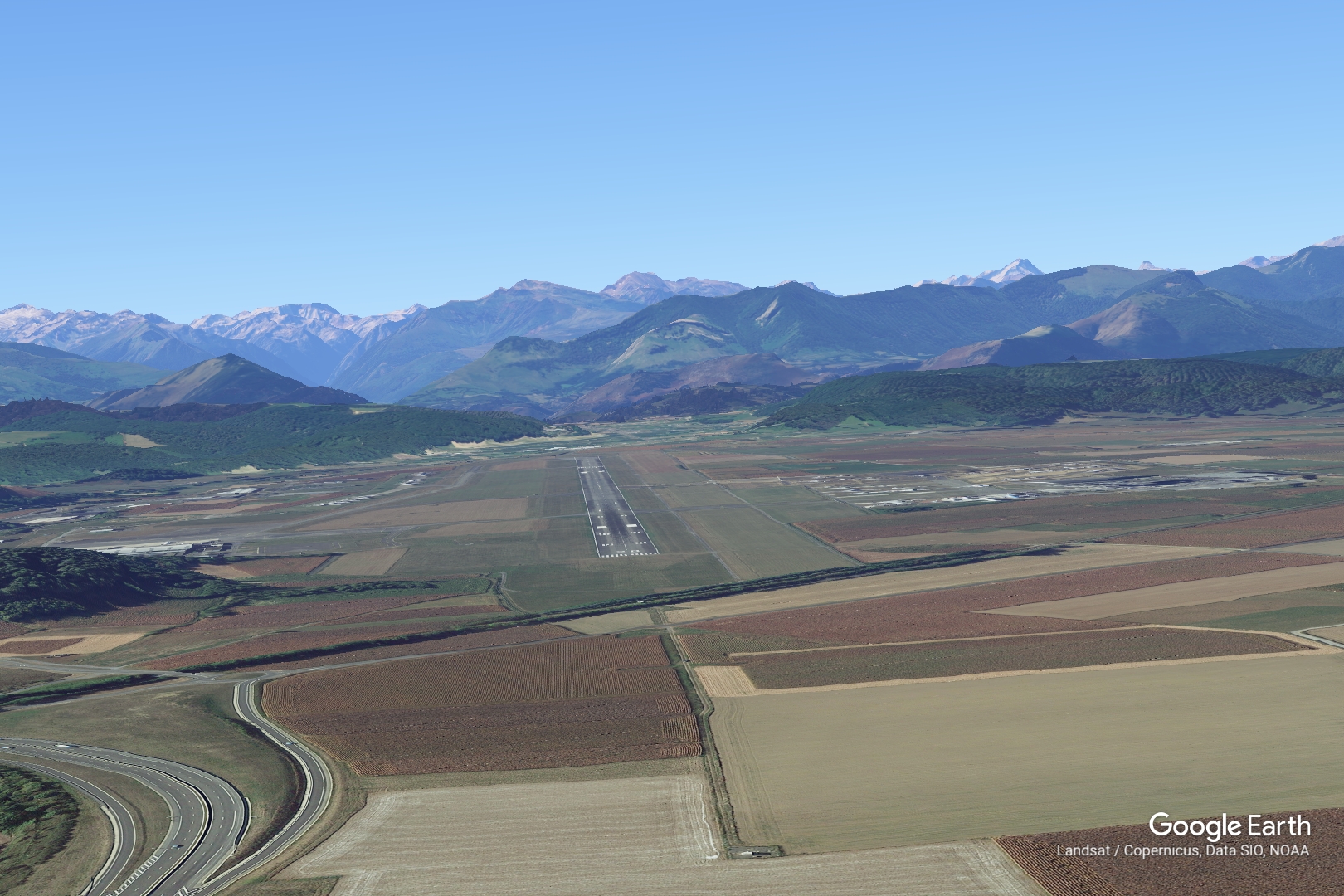}
    \end{subfigure}
    \caption{Illustration of the quality of the synthetic images - { Comparison of a real landing footage (left) with a synthetic replica (right)}}
    \label{fig:real_vs_synt_tarbes}
\end{figure}

As collecting synthetic data is easier than real ones, we opted to create a dataset primarily composed of synthetic images. In particular, we considered Google Earth Studio as a suitable choice because it provides high-quality runway images, as illustrated in Figure~\ref{fig:real_vs_synt_tarbes}. 
This figure compares an image recorded during a flight with our generator. Although the weather conditions differ between the two images, we note a great similarity in the runway's environment.

To fulfill the need for important volume of data, 
we chose to develop an open source data generator capable of reliable automatic annotation, 
and we decided to enrich the synthetic images that it can produce with images from real-footage. This, in turn, increases the overall dataset quality and allows to extend the detection task to a more realistic setting.
For that purpose, we collected videos of real landings (such as \footnote{\url{https://www.youtube.com/user/TheGreatFlyer}}) and requested the permission to use them for academic purpose. Once this authorization granted, we extracted some images of landing footage and performed manual runway annotation to enrich the dataset.
Since the process of labeling images is time-consuming, we included only 103 images in our dataset. However, videos can still be used to evaluate the performance of a model as long as a human reviews the output.

The remainder of this document is structured as follows.
Section~\ref{sec:related_work} presents the related works
on available datasets
and on vision-based runway detection approaches. 
Section~\ref{subsec:traj_def} presents the approach followed to design the dataset
and Section~\ref{sec:dataset} describes the \lard dataset. 
Section~\ref{sec:synth_generation} presents briefly the synthetic image generator.

\section{Related work}\label{sec:related_work}
Training Machine Learning models typically requires large amounts of data especially when tight performance guarantees are needed. Collecting and annotating this data is not only a costly task but also a particularly difficult one when all possible operating parameters need to be covered, including edge cases.

\subsection{Runway Images Datasets}
 To the best of our knowledge, there is no image dataset of runways seen from aerial front-view.
 Synthesized data is one solution to address this problem, as it allows the creation of multiple scenarios at a lower cost.  This is a well-known problem in aviation and the common solution is to use a flight simulator~\cite{lee2017flight}. For example, thanks to simulators, pilots can train for emergency maneuvers and become familiar with flight controls and standard procedures. The complexity and realism of such simulators depend on the needs and regulations of the relevant authorities (e.g. FAA in America, EASA in Europe~\cite{longridge2001simulator}).

To reproduce the environment around airport runways, our solution is to rely on an open-source virtual globe. Virtual globes are indeed prevalent tools in data collection, exploration, and modeling, used in numerous research activities over the last decade~\cite{yu2012google}. In this work, we consider Google Earth Studio~\cite{GoogleStudio}, an advanced animation tool for accessing and rendering Google satellite images. 

\subsection{Airport Data}\label{subsec:airport_database}
To generate a runway dataset, it is necessary to know the runway location.
The construction of a database with airports location is possible through the sharing of open source databases including geographic or thematic data. These include GeoNames~\cite{GNO}, established by a European organisation, the USGS Geographic Names Information System~\cite{GNIS}, established by the US government, and the GEOnet Name Server~\cite{GNS}, established by the US military. The thematic airport database includes OurAirports~\cite{OAP}. Complementary tabular information of airports and their runways can also be found in~\cite{cheng2017remote, yang2010bag, xia2018dota, xia2017aid} and have been used in previous works.
The previous databases can be merged to provide an approximate location of an airport runway. However, to our knowledge, there is no open source database that contains the location of the corners of a runway. Obtaining this data therefore requires manual annotation which can be applied either at the image level or in a geographic coordinate system using an online tool, such as Google Earth, and automatically projected into the image frame. The latter option, although more efficient, requires camera information. 
Those information can then be plugged into a remote sensing image provider such as Google Earth~\cite{GoogleEarth}, but limited to the available catalog of the platform.

\subsection{Vision-Based Runway Detection}
This section records the various efforts 
for the detection and localization of airport runways (not only based on deep learning).
The first line of work is based on image processing. The main focus consists in searching for runways’ singular features such as templates, geometric patterns, or textures. Those works are
 mainly based on pipelines using filtering operations such as Hough transforms, Sobel or Canny edge operators, and edge-preserving smoothing techniques~\cite{nazir2018vision,meng2006method,pi2003airport, aytekin2012texture, dong2011runway, sun2007edge}.
More recently, machine learning algorithms was applied, such as SVM or AdaBoost, mainly to classification tasks (airport versus non-airport)~\cite{budak2016efficient, cruz2013concrete, zhuang2014real, zhang2020runway, zongur2009airport}.

All these works suffer from multiple limitations. Firstly, image-processing-based methods considerably reduce the representation space and rely heavily on expert knowledge and a priori in the domain of operation such as~\cite{budak2016efficient}. Their robustness to outliers such as objects in the vicinity of the runway is limited, as is their generalization ability. Moreover, these algorithms require a large amount of data coupled with a lot of potentially complex and expensive metadata (\textit{airport design, photometric measurements, ...}).

The literature on airport runway detection and localization tasks using deep learning is scarce. Object detection and image classification tasks have largely benefited from deep learning-based methods. When it comes to the aeronautical domain, we note the low rate of publications, in particular on detection tasks. Some works are more or less related to the aeronautical domain, by integrating aircraft in their classification task. 
\cite{dhulipudi2021geospatial} have also proposed a deep learning approach for a related task which is the detection of ground markings on runways, while~\cite{koopman2022real} used a YOLO detector to detect airports from satellite images. The pioneering work in runway extraction is highlighted in ARDN~\cite{li2020research}. However, the impact of these works remains limited by the lack of reproducibility, as well as by the angle of view adopted for the detection. Indeed, the extraction of airports' runways is done by a satellite view from above. The input distribution is thus not compatible with runway detection from an airborne front-looking camera.
Today, to our knowledge, only private aeronautical companies challenge deep-learning-based solutions for runway detection with a front-looking camera either located on the nose or a wing of the aircraft. The success of deep learning for runway extraction has been notably demonstrated in recent research and development efforts of private companies such as Airbus’ Autonomous Taxi, Take-Off, and Landing demonstrators (ATTOL, Dragonfly,~\cite{au2022challenges}) or the Daedalean project.
Daedalean, in collaboration with the aviation certification authorities (FAA, EASA), published research reports tackling different certification and trustworthy challenges surrounding this use case~\cite{force2020concepts, balduzzi2021neural}. Neither the neural networks nor the training dataset are disclosed.

\section{Approach followed to design the dataset}\label{subsec:traj_def}
The \lard dataset is designed to represent civil aircraft landings.
Therefore we start by defining a \emph{generic landing approach cone} based on the documentation provided by aeronautical standards. Then, from this definition, we derive a strategy to generate a dataset of images with adequate labels.

 
 \subsection{Generic landing definition}
\begin{figure}[ht]
    \centering
    \includegraphics[width =.7\linewidth]{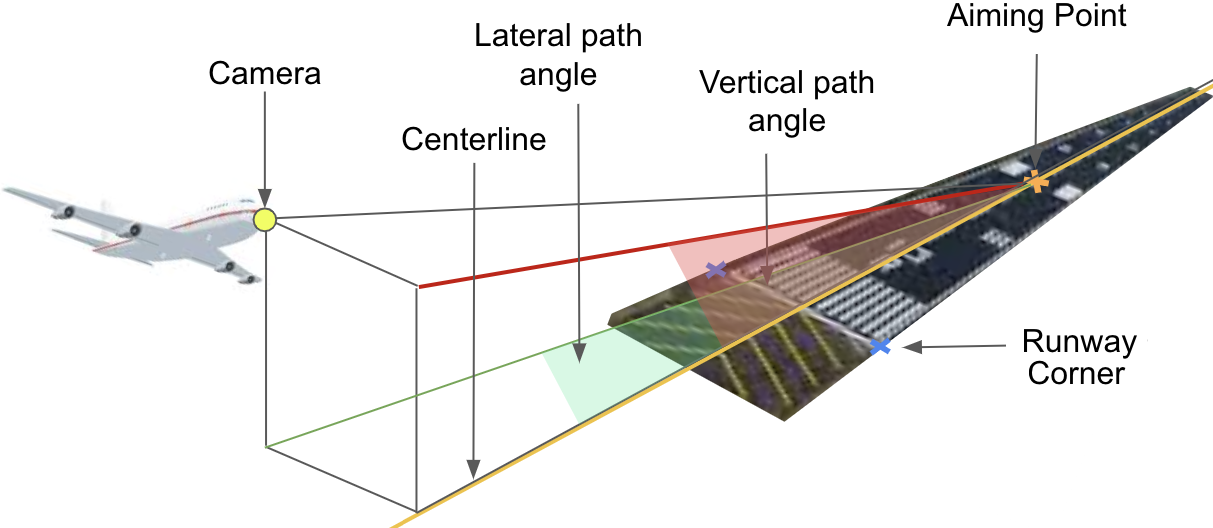}
    \caption{Geometry of a landing}
    \label{fig:camera_setting}
\end{figure}

 Figure~\ref{fig:camera_setting} illustrates 
 the different positions / angles / distances / markings involved in the geometric description of a landing. Runway markings are standardized~\cite{runway_marking} and appear in most cases as follows:
 A first line at the start of the runway, called \emph{landing threshold}, represents the underline limit of the runway. It is usually followed by a pattern of stripes (the \emph{piano}) and then the runway identifiers. The target of an aircraft during landing is the \emph{Aiming Point}, located 300 meters beyond the landing threshold, between two rectangular markings visible on each side of the runway \emph{centerline}\footnote{An imaginary line going through the middle of the runway}.

The position of the aircraft with respect to the runway is defined by 3 parameters:
The along track distance which corresponds to the distance between the projection of the aircraft nose on the centerline of the runway (on the ground) and the Aiming Point.
The lateral (resp. vertical) path angle which corresponds to the angle formed by the centerline and the line defined by the Aiming Point and the plane nose projection on the ground (resp. plane orthogonal to the ground going through the centerline).
On the other hand, the attitude of the aircraft is defined by its rotation angles 
(denoted respectively as pitch, roll, yaw). The yaw angle is relative to the runway heading\footnote{For instance a yaw of 0° indicates that the aircraft faces directly the runway, regardless of the runway orientation.} whereas pitch and roll are relative to the horizontal plane.

\begin{table}[t]
    \centering    
    \resizebox{.45\columnwidth}{!}{
        \begin{tabular}{|l|l|}
            \hline
            Parameter & range\\
            \hline
            \hline
            \texttt{Along track distance}          & {[}0.08, 3{]} NM    \\
            \texttt{Vertical path angle}    & [-2.2, -3.8]°  \\
            \texttt{Lateral path angle} & {[}- 4, 4{]} °     \\ 
            \texttt{Yaw}               & {[}-10,10{]} °   \\
            \texttt{Pitch}             & {[}-8,0{]} °     \\
            \texttt{Roll}              & {[}-10,10{]} °   \\ 
            \hline
        \end{tabular}
    }
    \captionof{table}{Parameters of the generic landing approach cone}
    \label{tab:odd_params}
\end{table}

These 6 parameters allow to define a generic landing approach cone (Definition~\ref{def:generic_approach_cone}) corresponding to a realistic aircraft trajectory during landing, as well as an envelope for the aircraft attitude that encompass typical aircraft orientations during approaches on a runway.

\begin{definition}[Generic landing approach cone] A generic landing approach cone is the set of all pairs $\langle$positions, attitude$\rangle$ within the ranges of the 6 parameters of Table~\ref{tab:odd_params}.
\end{definition}\label{def:generic_approach_cone}

 \subsection{Strategy to generate the dataset}\label{subsec:strategy_generation}
 Before designing the dataset, we first need to properly define the tasks it will address.
 \begin{mytask}[Main task]
 \label{task:main_task}
  The main task is the detection of a single runway within an image when the aircraft flies within the generic landing approach cone.
  We assume that the camera is positioned at the aircraft nose and directly faces the runway. 
Thus, the landing geometry defined previously directly applies to what can be observed from the camera.
For this task we chose to restrict the conditions to:
\begin{enumerate}[topsep=1pt,itemsep=0pt, partopsep=0pt, leftmargin=*]
  \item The aircraft is landing on airports with a \emph{piano}; 
  \item There exists only one runway for which current position is considered within the approach cone\footnote{Another runway can still be visible, but the aircraft should not be in its approach cone};
  \item The runway is fully visible on the image (no occlusion);
  \item Optimal conditions: clear daylight and no adverse weather conditions (clouds, precipitations...).
\end{enumerate}
\end{mytask}

An adequate dataset for this task~\ref{task:main_task} should therefore not only cover a variety of airports all around the world, but also span a wide range of positions inside the approach cone, to ensure a comprehensive coverage of all possible landing scenarios.
This motivated us to generate scenarios similar --in term of along track distances distribution-- to a complete landing approach, for several airports, and to produce few hundreds of pictures for each runway in the training set. The selection of this order of magnitude resulted from a trade-off between the variety of images produced for each runway, and the benefits in term of annotation cost reduction. 
Indeed, on one hand, the risk with having thousands of images per runway or more is the high similarity of resulting positions in the cone and the low independence between each image, which may lead to overfitting models. 
On the other hand, collecting only a few dozen of images per runway limits the possibility to encounter edge cases for each parameter and increases the need for manual annotation of runway corners to fulfill the high volume of data required. 
Finally, note that in practice, in an attempt to realistically cover the variety of possible landings, we generate approaches within the cone by adding Gaussian noise to the center of the ranges for each cone parameter.

Because large sets of data are needed, we chose to make use of synthetic images generators, and we selected Google Earth Studio for its availability and the configuration capabilities it offers. This tool support trajectories of positions (defined within our landing approach cone) as input, and allows to produce a variety of high quality images, relatively close to the reality, as illustrated in Figure~\ref{fig:real_vs_synt_tarbes}.
This choice led us to derive two tasks from the main task~\ref{task:main_task}.
The first one remains in the synthetic domain
whereas
the second addresses the Sim-to-Real capabilities.
\begin{mytask}[New runway generalization capacity]
 \label{task:generalization_task}
The task is the detection of runways never seen during training on synthetic images. 
\end{mytask}
To perform the task~\ref{task:generalization_task}, we provide, in the test set, synthetic runway images from a great variety of airports that were never seen in the training dataset.
\begin{mytask}[Sim to real generalization capacity]
 \label{task:sim_2_real_task}
 The task is the detection of runways on real footage images when training is done on synthetic images for the same runways. 
  \end{mytask}
To perform the task~\ref{task:sim_2_real_task}, we gathered and manually labelled frames from videos of landing available on the web.
It is worth noting that the pictures from real footage are in majority in 16:9 aspect ratios and 3840$\times$2160 or 1920$\times$1080 resolutions. In order not to lose information, we decided to keep each footage in its original resolution, thus having 3 different images sizes (the third one being 2448$\times$2648, which is the resolution chosen for the synthetic data).
Because real videos were captured from commercial aircraft, they are by definition in the generic landing approach cone. However in that case, the images had to be labeled manually a posteriori, by 
pointing the 4 corners of the polygons corresponding to the runways with a labeling tool.\\

\noindent\textbf{Extensibility} It should be emphasized that this dataset was also designed in such a way that users can easily extend it.
Indeed, we provide the set of airports from which the runways were taken, in the form of a database which contains the coordinates of runways corners and can be enriched with new airports and runways if needed. This can be done by using a dedicated script, where the user must indicate the geographic coordinates\footnote{Latitude, longitude and altitude} of each corner of a specific runway while specifying its metadata (the corresponding airport ICAO code and the runway identifier).
We also provide the possibility to generate new scenarios for Google Earth Studio and to benefit from an automatic labeling process.

\subsection{Choice of label associated to a runway} \label{subsec:label_selection}
As specified in task~\ref{task:main_task}, the images of the dataset must always contain fully visible runways.
Any label associated to an image should allow to define the runway inside of it in an unambiguous way whether the data is synthetic or real footage. There are several approaches for delimiting a runway, the most usual being contours, ground marking, corners, horizon line or any other semantics specific to a runway.
We chose to encode the runway position by the pixel coordinates of its four corners in the image. 

As pointed out in~\cite{balduzzi2021neural}, representing the runway by its four corners poses some concerns such as instability in the presence of runway occlusion and sensitivity to the aircraft position estimation.
The occlusion problem can be avoided by restricting the task to cases where there is no occlusion.
But the sensitivity issue still remains. While it was partially tackled by~\cite{balduzzi2021neural} by adding expert knowledge such as the width of the runway, we consider this stability problem to be out of scope of the open source dataset definition. 

However, the drawbacks mentioned previously were outweighed by the advantages of the corner representation, as this approach is easily applicable to any image to ensure consistency of metadata, and does not require camera-related information (typically camera angles). This allowed us to collect both synthetic images and real data from landing videos without any description of the camera.
This is especially true for real landing footage retrieved from Youtube which do not contain information on the runways or about the aircraft relative position.
Moreover, the detection of a runway by its four corners is a variant of box or parallelogram detection problems, which have been widely studied in deep learning~\cite{zaidi2022survey}. Thus, this literature can be reused to address the tasks identified in Section~\ref{subsec:strategy_generation}. 
Finally, it is compatible with both image detection and image segmentation approaches, two of the most widely used approaches for locating and identifying objects in image.

\section{Dataset}\label{sec:dataset}
This section presents in details the dataset that was designed according to the strategy described in Section~\ref{subsec:strategy_generation}. We also provide an assessment of its quality with respect to the generic landing.

\subsection{Data Description}
\label{subsec:data_description}
The dataset is divided into a  training set
and a test set
as illustrated in Figure~\ref{fig:repartition-test}.

\begin{figure}[hbt]
    \centering
    \begin{tikzpicture}[scale=0.086]
        \filldraw[turquoise] (0,0) rectangle (110.4,6);
        \filldraw[greenish] (110.4,0) rectangle (129.7,6);
        \filldraw[redish] (129.7,0) rectangle (150,6);
        \node[text width=1.5cm] at (68,2.8) {train};
        \node[text width=1.5cm] at (120,2.7) {train\_da};
        \node[text width=1.5cm] at (147,2.7) {test};
        \node[above, text width=1.5cm] at (66,6) {12~212};
        \node[above, text width=1.5cm] at (124,6) {2~221};
        \node[above, text width=1.5cm] at (145,6) {2~315};
        \node[below, text width=1.5cm] at (68,0) {73\%};
        \node[below, text width=1.5cm] at (125,0) {13\%};
        \node[below, text width=1.5cm] at (146,0) {14\%};
    \end{tikzpicture}
    \caption{Proportion of each subset of the dataset}
    \label{fig:repartition-test}
\end{figure}

\noindent The \textbf{Training set}
 is solely composed of synthetic images produced using the synthetic image generator described in Section~\ref{sec:synth_generation}. It is composed of 14~433 images of resolution 2448$\times$2648, taken from 32 runways in 16 different airports in total. It corresponds to approximately 451 pictures per approach (or per runway). It is worth noting that a subset of 5 runways in this training set are dedicated to Domain Adaptation (namely \emph{train\_da}), as described in the task~\ref{task:sim_2_real_task}.
\newline

\noindent The \textbf{Test set}
 aggregates two main sources of data and contains both synthetic pictures and frames taken from real footage of landing. 
It is composed of 2315 images divided as follows:

 

\begin{itemize}[leftmargin=*]
    \item \textbf{Synthetic}: 2~221 synthetic images taken from 79 runways in 40 different airports, which corresponds to approximately 28 pictures per approach. This part of the test set produces a variety of environments and airport types which have not been seen during training, and is aiming at verifying the generalisation capabilities of the detection models.  
    \item \textbf{Real}: 103 hand-labeled pictures from real landing footage on 38 runways in 36 different airports, usually obtained using an in-cockpit camera. This subset is further divided into nominal cases, edge case, and images dedicated to domain adaptation intended to verify the Sim-to-Real capabilities that a model may exhibit. The images provided in this very last subset correspond to the 5 runways also provided in the training dataset. 
\end{itemize}

\subsection{Dataset Quality Analysis}\label{subsec:quality_assessment}

\begin{figure}[hbt]  
    \begin{minipage}[t]{.44\linewidth}
        \includegraphics[width=\linewidth]{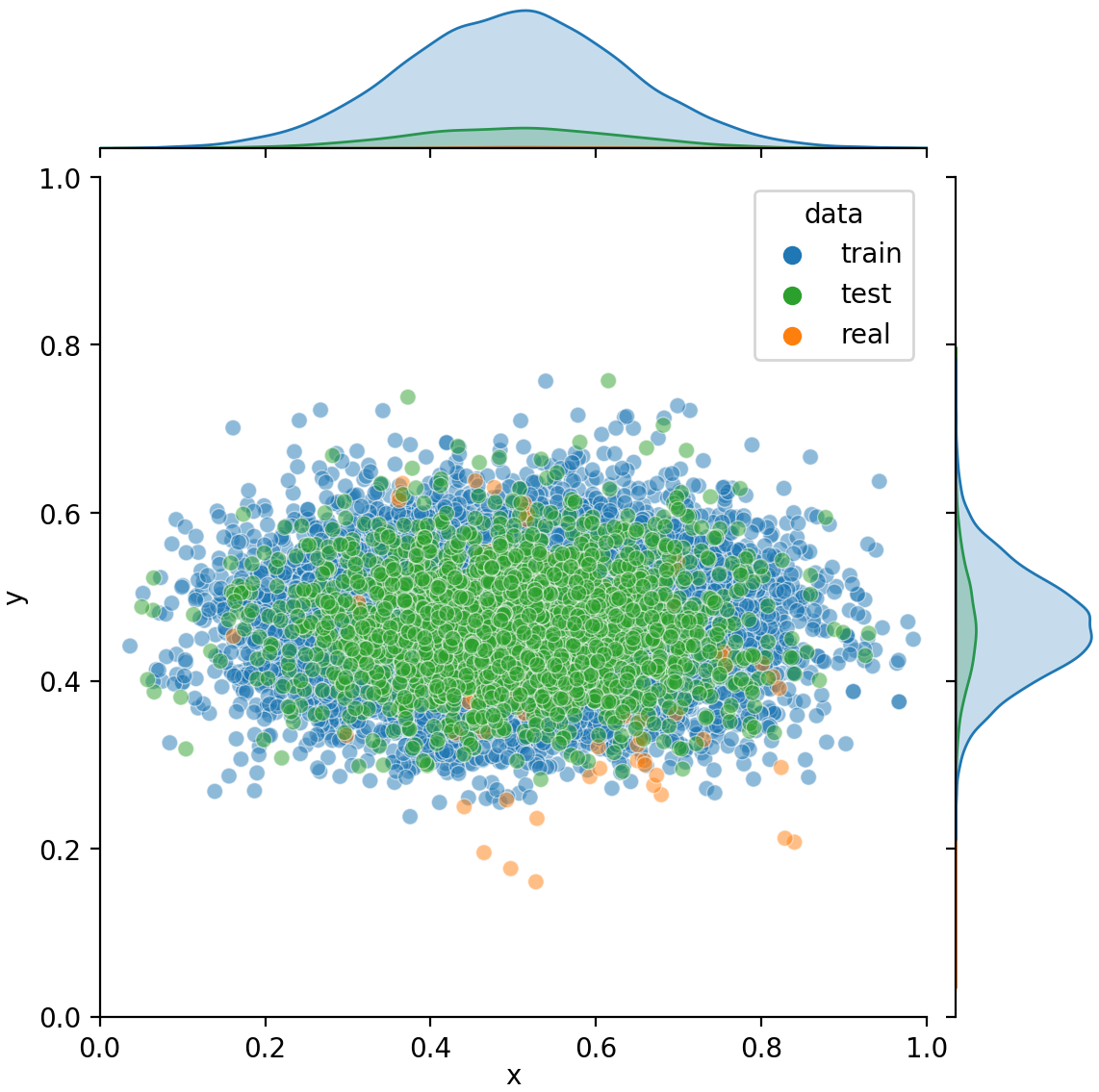}
    \caption[Normalized positions of runway centers in train and test and real subsets.]{Normalized positions of runway centers in train, test and real\protect\footnotemark~subsets.}
    \label{fig:runway_centers_positions}
    \end{minipage}
    \hfill
    \begin{minipage}[t]{.54\linewidth}
    \centering
    \includegraphics[width=\linewidth]{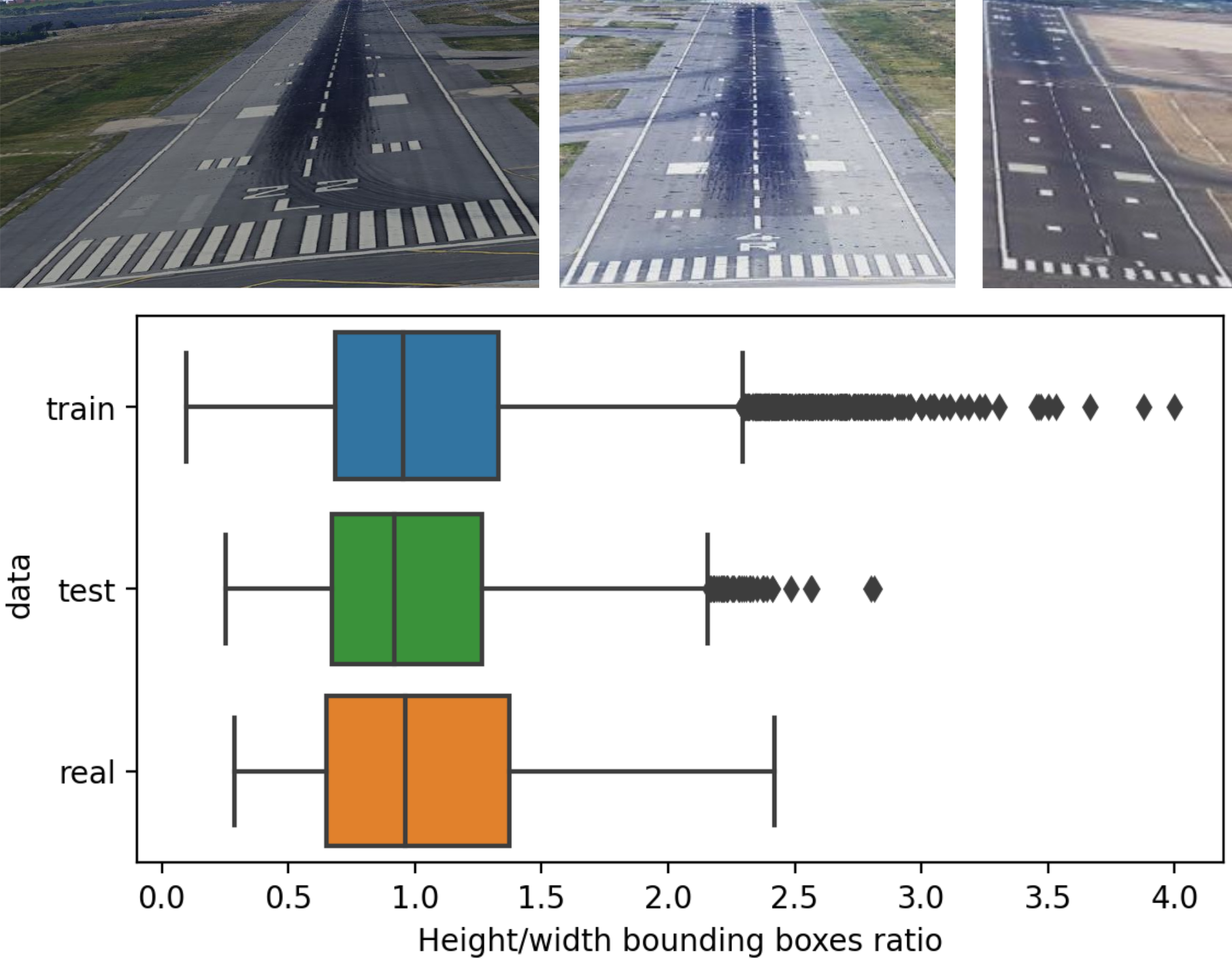}
    \caption{Top - \small{Illustration of different aspect ratios of the bounding boxes.} \normalsize{Bottom -} \small{Distributions of bounding boxes height over width ratios for the train, test and real subsets.}}
    \label{fig:bbox_ratio}
    \end{minipage}
 \end{figure}

\footnotetext{Subset of the test set containing only images from real footage.}

In this section, we provide a few statistical elements and the associated explanations to estimate the quality of the dataset and its suitability for the tasks presented in Section~\ref{subsec:strategy_generation}.
It covers the following claims:
\begin{enumerate*}[label=\arabic*)]
    \item The runways positions and aspect ratios of the bounding boxes which result from the image generation are homogeneously distributed and suitable for a detection task,
    and 
    \item the distribution of airports used to generate synthetic images is relevant to the tasks and produces a diversity of runway characteristics and surrounding terrains and landscapes.
\end{enumerate*}

\subsubsection{Runway positions}

The plot of runway centers positions of Figure~\ref{fig:runway_centers_positions} shows an even distribution both for the training set and for the test set, located primarily around the center of the images. Nevertheless, a large area in the top and the bottom contain little to no points, which is the result of two main factors: 
\begin{enumerate*}[label=(\roman*)]
    \item the presence of the watermark, which is expected to be removed from the images before usage by cropping 300 pixels from the top and the bottom of the pictures, and 
    \item the ranges of the \emph{pitch} parameter defined in the Table~\ref{tab:odd_params} which 
    prevent the runway to appear at the very top or bottom of the image.
\end{enumerate*}
Additionally, the real images of the test set appear to be slightly biased towards the bottom-right, which seems to result from the positions of the cameras in the cockpits.  

\subsubsection{Bounding boxes analysis}

\begin{figure}[hbt]
    \centering
    \begin{subfigure}[t]{.54\linewidth}
        \centering
        \includegraphics[width=\linewidth]{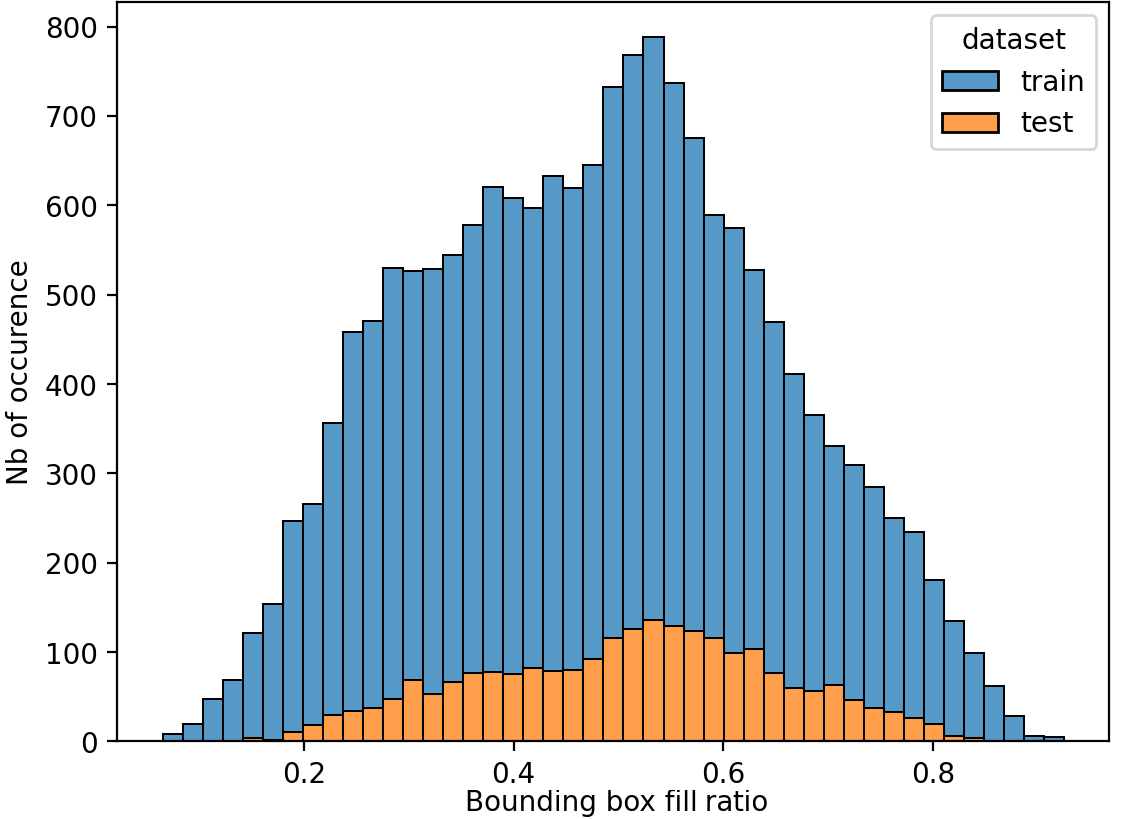}
        \caption{\footnotesize Distribution of bounding box fill ratios (percentage of the bounding box that correspond to pixels belonging to the runway itself)}
        \label{fig:bbox_filling_percentage}
    \end{subfigure}
    \hfill
    \begin{subfigure}[t]{.43\linewidth}
        \centering
        \includegraphics[width=\linewidth]{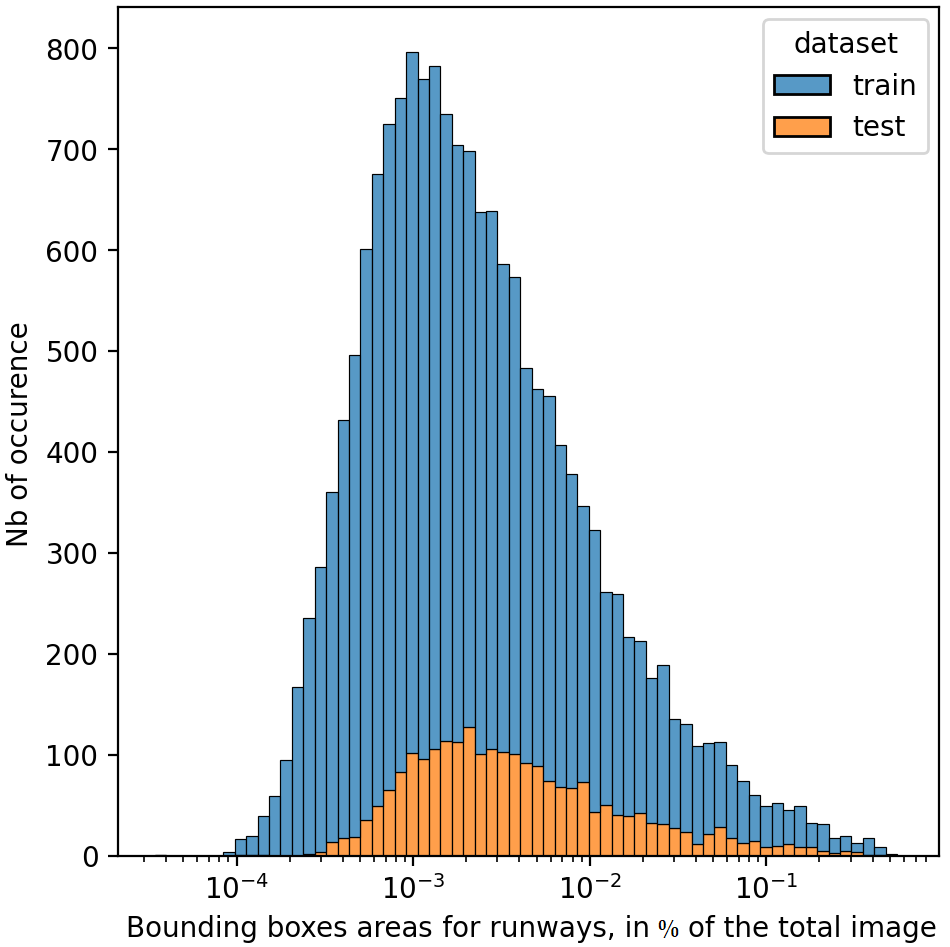}
        \caption{\footnotesize Distribution of bounding boxes areas (areas in logarithmic scale)}
        \label{fig:bbox_areas_percentage}
    \end{subfigure}
    \caption{Comparison of bounding box characteristics between training and test sets}
    \label{fig:histograms_bbox_comparison}
\end{figure}

The aspect ratio of the objects bounding boxes is a sensitive aspect for a detection task, as elongated objects in one or the other direction may not exhibit recognizable features. Figure~\ref{fig:bbox_ratio} illustrates the aspect ratio variability, and highlights how the majority of the bounding boxes in all three subsets have an aspect ratio between 0.5 and 1.5, indicating that most images are suitable for the targeted detection task.

The histograms of Figure~\ref{fig:histograms_bbox_comparison} illustrate the relationships between the runways, their bounding boxes and the global images. Figure~\ref{fig:bbox_filling_percentage} shows comparable distribution for the training and the test set, where most of the runways fill between 20$\%$ and 80$\%$ of their bounding boxes. This also indicates that bounding boxes should in general contain enough runways pixels for the detection task to be applicable and consistent\footnote{Note that for the labels, we use the ground truth directly from a geometric projection, therefore the entirety of the runway polygon is inside of the bounding box}. 
Additionally, Figure~\ref{fig:bbox_areas_percentage}, which illustrates how the areas of the bounding boxes cover the whole images, shows that the training set and the test set follow approximately the same distribution. 
This provides a certain level of guarantee that the bounding boxes will look similar between the training and the test set. 
Moreover, the figure shows that the vast majority of bounding boxes 
areas are over $25\times25$ pixels, which makes them large enough
 for a runway to be detected by humans.
On the other hand, the dataset contains only a few examples of bounding boxes with large size, which may bias the learning process when the aircraft is close to the runway 
and should be further investigated.

\subsubsection{Distances to runways}

\begin{figure}[hbt]
    \centering
    \vspace{-10pt}
    \includegraphics[width=0.42\textwidth]{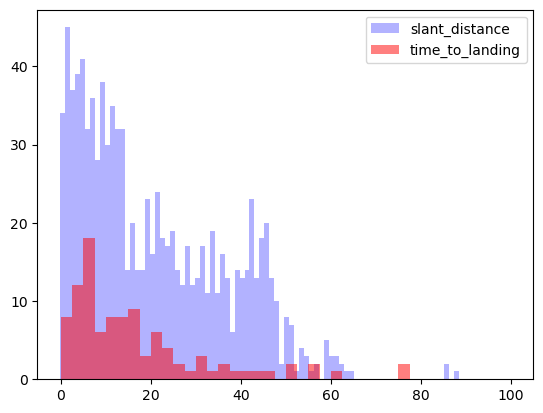}
    \caption{Comparison of distance estimation between real images and synthetic images in the test set}
    \label{fig:slant_distance_ttl}
\end{figure}

The synthetic images and the real images do not contain the same metadata. The distance between the aircraft and the runway is given for synthetic images as the \emph{slant distance}, however it is not available for real images, for which a value called \emph{time to landing} is provided instead. This value can be used as a proxy for the distance to the runway, considering that planes have comparable speed during landing phase.

Figure~\ref{fig:slant_distance_ttl} shows how the distributions of \emph{slant distance} (for synthetic images) and  \emph{time to landing} (for real images) relate to each other\footnote{Only the shapes of the distributions should be compared as the \emph{slant distance} was re-scaled to fit the diagram}. It indicates that for both sources of data, the test set contains an important part of the images close to the runway while a non-negligible number of pictures were taken at longer distances from the runway, in a nearly evenly distributed manner, despite the limited number of real images.

\subsubsection{Visualisation of the approach cone}

\begin{figure}[ht]
    \centering
    \begin{subfigure}{0.48\textwidth}
        \includegraphics[width=\textwidth]{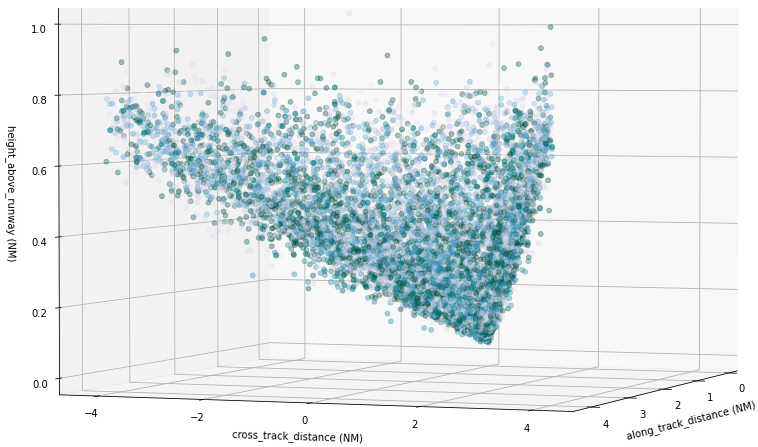}
        \caption{ \footnotesize Training set}
        \label{fig:trajectories_of_landings_train}
    \end{subfigure}
    \hfill
    \begin{subfigure}{0.48\textwidth}
        \includegraphics[width=\textwidth]{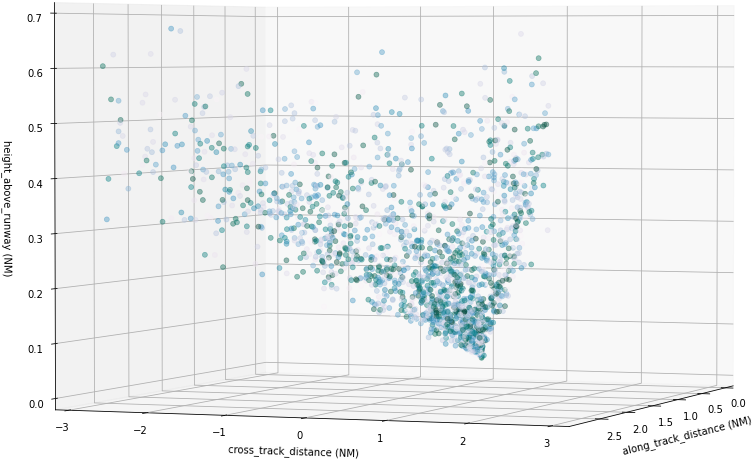}
        \caption{\footnotesize Synthetic test set}
        \label{fig:trajectories_of_landings_test}
    \end{subfigure}
    \caption{3-dimensional visualisation of aircraft positions in the training set and the test set}
    \label{fig:trajectories_of_landings}
\end{figure}


The parameters used in the scenario generation correspond to the standard approach cone of an aircraft during landing phase (Definition~\ref{def:generic_approach_cone}). Figure~\ref{fig:trajectories_of_landings} illustrates the distribution of points in this cone for synthetic images, for which the position of the aircraft relative to the runway can be retrieved. In this figure, the z-axis correspond to the \textit{along track distance}, but the other two axis are also distances (\textit{cross track distance} and \textit{height above runway}), computed from the angles provided in Table~\ref{tab:odd_params} (\textit{Lateral path angle} and \textit{Vertical path angle}). For the training set in Figure~\ref{fig:trajectories_of_landings_train}, the randomly sampled points span the whole approach cone corresponding to the scenarios generation parameters. Moreover, while the synthetic test set contains less data, it still covers a variety of positions in the cone, as illustrated in Figure~\ref{fig:trajectories_of_landings_test}.

\subsubsection{Airport worldwide distribution}

\begin{figure}[t]
    \centering
    \includegraphics[width=.8\linewidth]{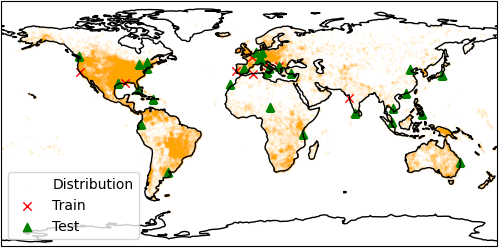}
    \caption{Distribution of airports used for the training set and the test set}
    \label{fig:airport_distribution}
\end{figure}

Figure~\ref{fig:airport_distribution} plots the distribution of airports from all around the world which were used to build the \lard dataset. Indeed, obtaining a great variety of images is a fundamental aspect for verifying the generalization capabilities of the models, as highlighted in task~\ref{task:generalization_task}, and current distribution of airports presents the following benefits:
first, it ensures a diversity of runway visuals, with different surface types\footnote{Asphalt and concrete are typically used for runway surfaces} and  various runway length, width and markings, even if the runway standardization reduces the variability for this aspect. Second, it allows for a variety of surrounding terrain and landscapes such as grass, snow, dirt, but also city architectures, water bodies or mountainous reliefs.

\section{Synthetic Dataset Generation}
\label{sec:synth_generation}
This section presents the synthetic image generator based on Google Earth Studio.
The generator does not only generate synthetic images but also relevant information associated 
to each image (e.g. runway position, position of the camera).

\subsection{General overview of Google Earth Studio}
Google Earth Studio allows to draw trajectories or zoom from one point to another and render the corresponding synthetic pictures or videos. For partial automation, the tool supports scenario files in \textit{.esp} or \textit{.kml} formats containing the parameters for sequences of frames. It is thus possible to generate and render a synthetic set of pictures from a \textit{.esp} file where each frame of the video obtained can then be associated with metadata. In Table~\ref{tab:earth_parameters} we list some of the parameters that are configurable by the user.

Although the generated data are based on real satellite images, the underlying transformations for the aggregation of satellite images and their adaptation to user constraints (day, night, time, cloudy, see Table~\ref{tab:earth_parameters}), are not disclosed and may potentially induce synthetic biases.

\begin{table}[hbt]
    \centering
        \begin{tabular}{|l|l|l|}
            \hline
            \multirow{3}{*}{Camera} & Position & Longitude, Latitude, Altitude \\ \cline{2-3} 
             & Rotation & Horizontal angle, Vertical angle, Roll \\ \cline{2-3} 
             & Field of view &  \\ \hline
            Environment & Date & Time of year, month, day, hour \\ \hline
            \multirow{5}{*}{Rendering} & Output type & Pictures (\textit{.jpeg}) or video (\textit{.mp4}) \\ \cline{2-3} 
             & Dimensions & Width, height of the output \\ \cline{2-3} 
             & Coordinates & Metadata with 3D positions (\textit{.json}) \\ \cline{2-3} 
             & Texture & Image quality \\ \cline{2-3} 
             & Attribution position & Position of the image attribution \\ \hline
        \end{tabular}
    \caption{Earth Studio parameters}
    \label{tab:earth_parameters}
\end{table}

\subsection{Generator overview}
\label{subsec:earth_query}
\begin{figure}[hbt] 
\centering 
    \resizebox{.95\linewidth}{!}{%
        \begin{tikzpicture}[thick,scale=0.5]
\usetikzlibrary{shapes,shapes.geometric,calc}
\usetikzlibrary[shadows]

\tikzstyle{block} = [draw,minimum height=2em, minimum width=1.5cm, inner sep=3pt];
\tikzstyle{txt} = [text centered, inner sep=0pt];
\tikzstyle{compliance} = [dashed, draw,  inner sep=3pt ];
\tikzstyle{database} = [cylinder, 
        shape border rotate=90, 
        draw,
        minimum height=1cm,
        minimum width=2cm,
        shape aspect=.25,];

\tikzset{
  multidocument/.style={
    shape=tape,
    draw,
    fill=white,
    tape bend top=none,
    double copy shadow},
  manual input/.style={
    shape=trapezium,
    draw,
    shape border rotate=90,
    trapezium left angle=90,
    trapezium right angle=80}}

\path (0,-2) node[database, fill= lightgray] (airportbase) {
\begin{tabular}{c} runway\\ database \end{tabular}
};

\path (airportbase)+(0,4) node[block, fill= lightgray, dashed] (confi) {
\begin{tabular}{c} Configuration\\ file \end{tabular}
};

\path (airportbase)+(7,2) node[block] (scengen) {
\begin{tabular}{c} Scenario\\generation \end{tabular}
};

\path (scengen)+(5,0) node[block] (ges) {
\begin{tabular}{c} Google\\ Earth\\Studio \end{tabular}
};

\path (ges)+(4.5,0) node[block] (label) {
Labelling
};

\draw  [->]   (airportbase.east)  -| ($(airportbase.east)+(1,2)$) -- (scengen.west);

\draw  [->]   (confi.east)  -| ($(confi.east)+(1,-2)$) --(scengen.west);
\draw  [->]   (scengen.east)  -- (ges.west);
\draw  [->]   (ges.east)  -- (label.west);

\path (label)+ (8.5,0) node[database, fill= lightgray,minimum height=2.5cm,minimum width=5.5cm] (res) {
};
\path (res)+(0,2) node (text) {
Synthetic images
};

\draw  [->]   (label.east)  -- (res.west);

\path (res) + (-3,0) node [multidocument] (img) {image.jpg};

\path (img) + (5,0) node [multidocument] (lab) {\begin{tabular}{c} labels \\+ metadata\end{tabular}};

\end{tikzpicture}
    }
\caption{Generator pipeline} 
\label{fig-gen-pipeline} 
\end{figure}
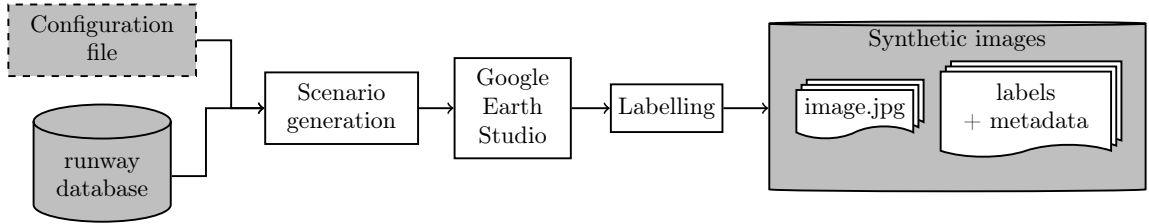
The generator pipeline is presented in Figure \ref{fig-gen-pipeline}.
The two inputs (in gray) are the airport database and the configuration file to be filled by the user, setting which runway they want to generate images from and other parameters (e.g. number of images). Then, the first script (in white) generates a scenario file that can be provided as an input for Google Earth Studio. 
This virtual globe tool can then generate the corresponding images, together with an information file  (here in json format). Finally, the last module of our generator associates the '\textit{labels}' to each image, in particular the scaled position of the four corners on the picture.

\begin{figure}[ht]
     \centering
     \begin{subfigure}[b]{.49\linewidth}
         \centering
         \includegraphics[width=.9\linewidth]{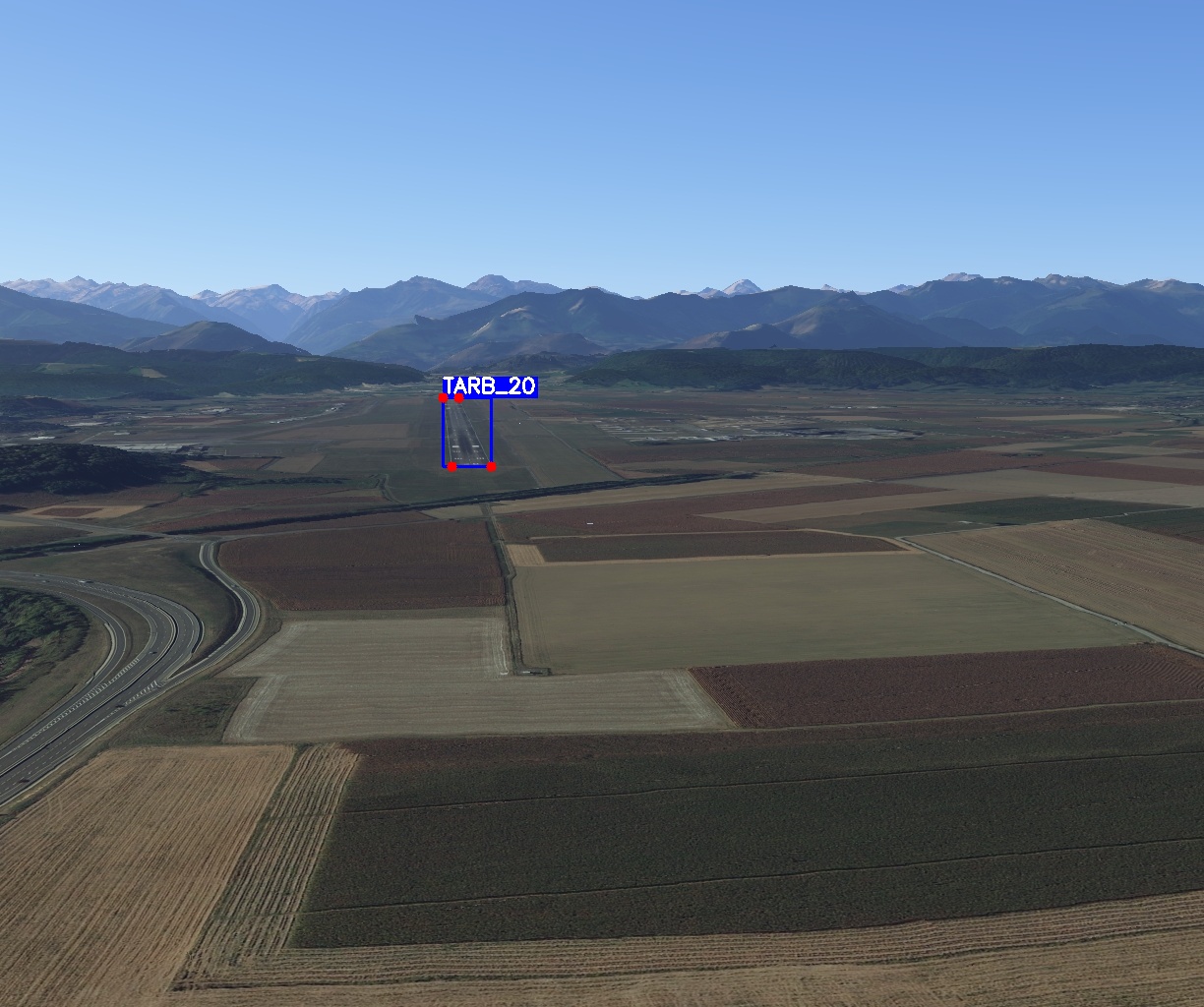}
     \end{subfigure}
     \hfill
     \begin{subfigure}[b]{.49\linewidth}
         \centering
    \includegraphics[width=0.9\linewidth]{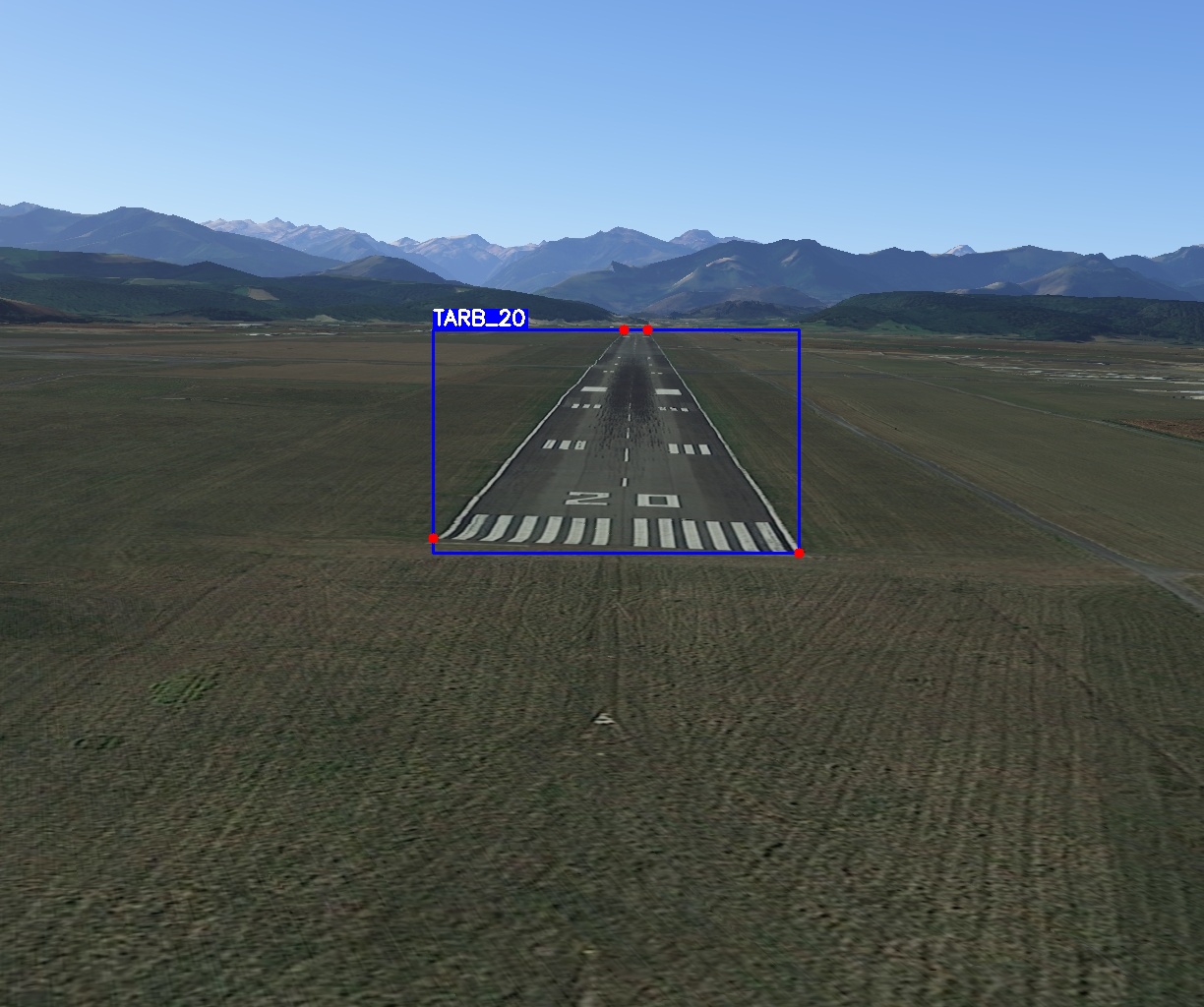}
     \end{subfigure}
        \caption{Images from Tarbes runway (France), with bounding boxes isolating the runway - Using the projection matrices, 
        we can project the coordinates of the runway corners onto any generated image.}
        \label{fig:Tarbes results}
\end{figure}

The output in gray contains the images, the labels and the metadata.
 This will be used as a ground-truth for benchmarking machine learning models for the tasks defined in Section~\ref{subsec:strategy_generation}. These labels allow for the synthesis of images like the ones shown in Figure~\ref{fig:Tarbes results}.

\subsection{Automatic Annotation}
This section details the labelling module of Figure \ref{fig-gen-pipeline}.
To generate the labels, we retrieve the positions of the corners and the position of the camera in the World Geodesic System 1984 (WGS84\footnote{Coordinate system for spatial referencing, navigation and cartography.}) coordinate system and we project the corners of the runway in the image coordinate system. 
Note that the \emph{aiming point} position can be deduced
in the world reference coordinate system WGS84, 
as long as the latitude and longitude coordinates of the runway corners are known. The projection from WGS84 to the image based coordinate system is done using two standard matrices \cite{szeliski2022computer}:
\begin{itemize}[leftmargin=*]
    \item The Extrinsic matrix whose role is to get the coordinates of the corners in the camera-centered coordinate system.
    \item The Intrinsic matrix whose role is to project the 3D coordinates expressed in the camera-centered coordinate system into the 2D image
\end{itemize}

\begin{figure}[t]
         \centering
         \includegraphics[width=.8\linewidth]{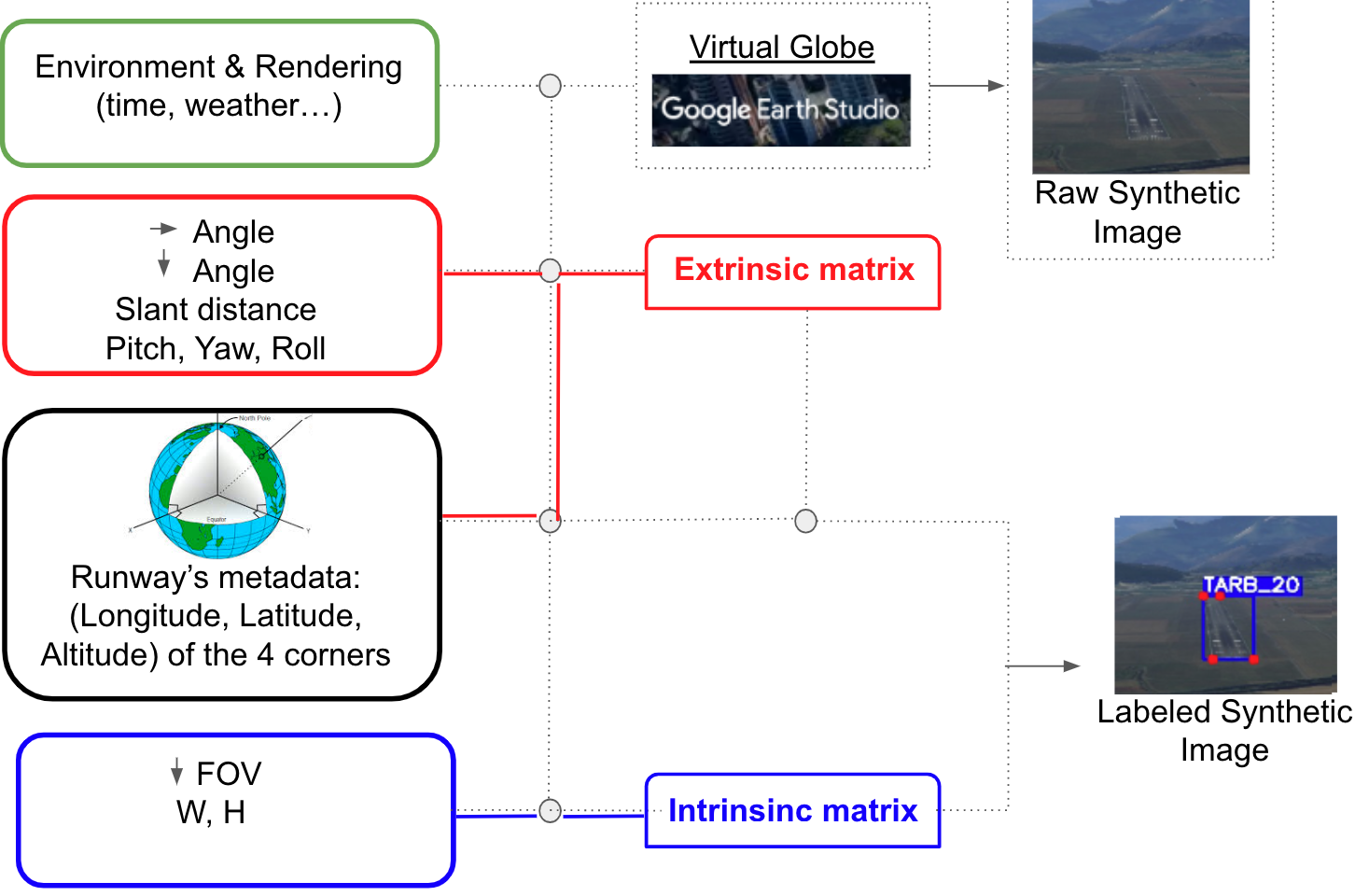}
         \caption{Synthetic annotation  pipeline}
         \label{fig:pipeline}
\end{figure}

The extrinsic matrix takes as parameters the rotation matrix of the camera as well as a translation vector. The rotation matrix can be easily deduced from the composition of the rotation along the three axis that depend on the pitch, roll and yaw angles. The translation vector only depends on the camera position which is determined by the \textit{aiming point}, the slant distance and the horizontal and vertical angles. 
The intrinsic matrix takes as parameters all the information related to the scale, field of view and the optical center of the camera. We instantiate the intrinsic matrix for the projection from the camera system to the image system, as proposed in \cite{hartley2003multiple}, using the vertical and horizontal focus values and the principal offset point. We can then deduce the focal length values using the dimensions of the image and the field of view, whose values are specified by the user during the generation of synthetic images.
The main steps of our synthetic annotation pipeline are described in Figure~\ref{fig:pipeline}. 

Overall, for the cost of a single annotation (the corners of a runway), this generator allows to produce an infinity of images with various camera angles and positions, where the annotation is automatically propagated, which drastically reduces the labeling cost.

\section{Conclusion and way forward}\label{sec:goals_scope}
According to EASA and \cite{balduzzi2021neural}, a majority of non-commercial airplane accidents occur during landing phase in good weather conditions. Moreover, they are often due to human errors, with perception being the highest risk factor, which highlights the need for safer landing systems.

Introducing autonomy in these systems could be a first step to solve this issue, starting with pilot assistance, up to fully autonomous landings in the far future. This evolution will undoubtedly rely on Artificial Intelligence to detect the runway position which, in turn, will allow to compute the aircraft position. We presented in this article an unambiguous specification of this task and we underscored the importance of large dataset to enable the use of deep learning algorithm. 

This paper is a first answer to the problem of collecting high volumes of aerial images. In the context of autonomous landing, we presented both a dataset of runway images and the synthetic generator based on Google Earth that allows to generate such images. 
A major benefit of this approach is the possibility to enrich the dataset at will with new runways, without incurring high annotation costs, thanks to the automatic labeling process. Thus we hope that this work will inspire the expansion of the dataset and serve as a foundation for future research in the wider context of object detection in aerial images.
Indeed, the parameters ranges used 
in this paper 
can be modified to suit any task relying on the production of aerial images, while
benefiting from the automatic labeling capabilities. 


\section{Acknowledgement}
This work has benefited from the AI Interdisciplinary Institute ANITI, which is funded by the French ``Investing for the Future – PIA3'' program under the Grant agreement ANR-19-P3IA-0004. The authors gratefully acknowledge the support of the DEEL \footnote{\url{https://www.deel.ai/}}
and PHYDIAS 2 projects.

\bibliographystyle{alpha}
\bibliography{bib}

\end{document}